\documentclass{article} 
\usepackage{spconf,amsmath,graphicx}

\usepackage{cite}
\usepackage{booktabs} 
\usepackage{url}
\usepackage{multirow} 
\usepackage{hyperref}
\usepackage{xcolor} 
\usepackage{amsfonts}

\title{LATINX: ALIGNING A MULTILINGUAL TTS MODEL WITH DIRECT PREFERENCE OPTIMIZATION}
\name{Luís Felipe Chary and Miguel Arjona Ramírez}
\address{Escola Politécnica, Universidade de São Paulo \\
         São Paulo, Brazil \\
         \small{\texttt{\{luisfchary, maramire\}@usp.br}}}

\begin{document}
\ninept 
\maketitle
\begin{abstract}
We present LatinX, a multilingual text-to-speech (TTS) model for cascaded speech-to-speech translation that preserves the source speaker’s identity across languages. LatinX is a 12-layer decoder-only Transformer trained in three stages: (i) pre-training for text-to-audio mapping, (ii) supervised fine-tuning for zero-shot voice cloning, and (iii) alignment with Direct Preference Optimization (DPO) using automatically labeled pairs based on Word Error Rate (WER) and speaker-similarity metrics. Trained on English and Romance languages with emphasis on Portuguese, LatinX with DPO consistently reduces WER and improves objective similarity over the fine-tuned baseline. Human evaluations further indicate stronger perceived speaker similarity than a strong baseline (XTTSv2), revealing gaps between objective and subjective measures. We provide cross-lingual analyses and discuss balanced preference signals and lower-latency architectures as future work.
\end{abstract}
\section{Introduction}
\label{sec:intro}

Despite increasing global connectivity, language barriers remain a significant challenge, particularly in regions like Brazil where fluency in English is limited \cite{b18}. While text translation tools have advanced, they fail to convey the rich paralinguistic information embedded in speech, such as emotion, tone, and vocal identity. State-of-the-art speech-to-speech (S2S) translation systems aim to bridge this gap, but preserving the original speaker's voice across languages remains a complex task.

Recent advancements in generative audio, particularly autoregressive models like VALL-E \cite{b16} that operate on discrete audio tokens, have shown remarkable capabilities in zero-shot Text-to-Speech (TTS). This paradigm, where a model clones a speaker's voice from a short audio prompt, provides a powerful foundation for creating more natural and personalized S2S systems.

This work builds upon that foundation by developing and analyzing \textbf{LatinX}\footnote{Audio samples are available at: \url{https://seu-usuario.github.io/latinx-demo}}, the core TTS component of a cascaded pipeline for multilingual, voice-preserving speech-to-speech translation. The core of our system is a decoder-only Transformer model trained in a three-stage process. The key contributions of this paper are:

\begin{itemize}
    \item The development of a complete, modular pipeline for multilingual TTS, featuring custom G2P and neural codec components;
    \item A detailed methodology for aligning the generative model using Direct Preference Optimization (DPO) \cite{b17};
    \item An extensive evaluation across six languages (en, es, fr, it, pt, ro), including a granular cross-lingual analysis;
    \item A discussion on the notable divergence between objective and subjective evaluation metrics for speaker similarity.
\end{itemize}


\section{Data and Pre-processing}
\label{sec:data}
The training corpus was compiled from three sources: Common Voice (v16.1, Dec. 2024) \cite{b1}, Multilingual Librispeech \cite{b2}, and a private collection assembled by the authors to increase the representation of Portuguese. We used data for six languages: English (en), Spanish (es), French (fr), Italian (it), Portuguese (pt), and Romanian (ro).

From an initial set of over 16,500 hours, we applied a strict filtering pipeline to ensure data quality. First, transcriptions containing non-phonetic characters or artifacts were removed. Second, audio quality was assessed using TorchAudio-Squim \cite{b3}, and we retained only samples meeting the following criteria: Scale-Invariant Signal-to-Distortion Ratio (SI-SDR) $\ge 10$, Perceptual Evaluation of Speech Quality (PESQ) $\ge 2.5$, and Short-Time Objective Intelligibility (STOI) $\ge 0.8$.

The final filtered dataset comprises approximately 9,736 hours of audio from over 84,000 unique speakers. This dataset is intentionally unbalanced, with Portuguese accounting for roughly 3,844 hours (approx. 40\% of the total), to foster a strong base model for this language. The linguistic imbalance is addressed during the fine-tuning stage via sub-sampling techniques. All audio files were resampled to 16 kHz. For the Grapheme-to-Phoneme (G2P) conversion, we utilized phonetic dictionaries from the `ipa-dict` repository \cite{b4}. A summary of the data composition before and after filtering is presented in Table \ref{tab:data_summary}.



\begin{table}[t]
\centering
\caption{Training data composition by language, showing total duration and number of unique speakers after quality filtering.}
\label{tab:data_summary}
\resizebox{\columnwidth}{!}{%
\begin{tabular}{lrrrr}
\toprule
\multirow{2}{*}{\textbf{Language}} & \multicolumn{2}{c}{\textbf{Duration (hours)}} & \multicolumn{2}{c}{\textbf{Speakers}} \\
\cmidrule(lr){2-3} \cmidrule(lr){4-5}
& \textbf{Original} & \textbf{Filtered} & \textbf{Original} & \textbf{Filtered} \\
\midrule
English (en) & 3,490.4 & 2,071.8 & 70,258 & 53,490 \\
Portuguese (pt) & 7,731.5 & 3,844.1 & 2,991 & 2,386 \\
French (fr) & 2,143.7 & 1,643.8 & 16,173 & 12,237 \\
Spanish (es) & 2,174.9 & 1,494.2 & 16,381 & 11,371 \\
Italian (it) & 963.6 & 663.6 & 6,580 & 4,936 \\
Romanian (ro) & 25.1 & 18.3 & 291 & 268 \\
\midrule
\textbf{Total} & \textbf{16,529.2} & \textbf{9,735.7} & \textbf{112,674} & \textbf{84,688} \\
\bottomrule
\end{tabular}%
}
\end{table}

\section{Model Architecture}
LatinX is a modular pipeline composed of three main components.

\subsection{Grapheme-to-Phoneme (G2P) Model}
To provide a robust phonetic input representation, we use \textit{LatPhon} \cite{b5}, a lightweight Transformer-based encoder-decoder model specifically trained for Romance languages and English.

\subsection{Neural Audio Codec and Vocoder}
We employ a neural audio codec, \textit{Spectrogram Patch Codec} \cite{b6}, to discretize speech. Its VQ-VAE component \cite{b7} transforms mel-spectrograms into a sequence of discrete indices from a codebook of 4096 vectors, using an encoder with a downsampling factor of 4 and a latent dimension of 512. The model is trained with a combination of a reconstruction loss ($l_1$ + LPIPS), an adversarial loss using a PatchGAN discriminator, and a vector quantization commitment loss. A HiFi-GAN vocoder \cite{b8} then converts the reconstructed mel-spectrograms back into a high-fidelity waveform.

\subsection{Autoregressive Generation Model: LatinX}
The core of our system is a 12-layer decoder-only Transformer with 210M parameters, inspired by LLaMA 2 \cite{b9}. The architecture features an embedding dimension of 1024, a 4096-dimensional feed-forward network, and 16 attention heads. For efficiency, we incorporate Rotary Position Embeddings (RoPE) \cite{b10} and FlashAttention \cite{b11}.

The model operates on a fixed context window of 8192 tokens, partitioned into three segments: 256 tokens for the input phonemes, 3968 tokens for the acoustic prompt (from a 3-second voice sample), and 3968 tokens for the autoregressive generation. It is trained to predict the subsequent audio token based on this combined context. On an NVIDIA RTX 4090 GPU with float16 precision, the model achieves an inference speed of approximately 130 tokens per second. Given that our neural codec produces roughly 630 tokens per second of audio, this corresponds to a Real-Time Factor (RTF) of approximately 4.85, making it suitable for offline synthesis.

Unless otherwise stated, we use repetition-aware sampling \cite{b20} with top-$p{=}1.0$,  temperature $T{=}0.7$, repetition window size $W{=}240$ and max repetition $M{=}10$. 

\section{Training and Optimization}
\begin{figure*}[t]
    \centering
    \includegraphics[width=0.8\textwidth]{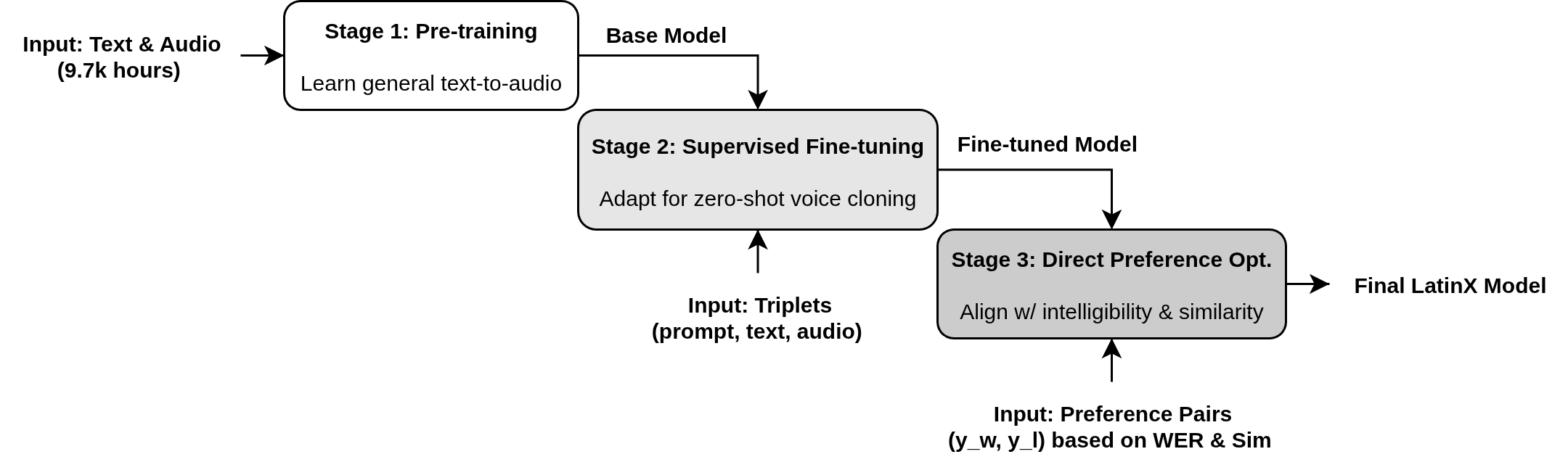}
    \caption{The three-stage training pipeline for the LatinX model, progressing from general pre-training to supervised fine-tuning for voice cloning, and concluding with DPO alignment.}
    \label{fig:training_pipeline}
\end{figure*}

The LatinX model is trained in three sequential stages (fig. \ref{fig:training_pipeline}) using the AdamW optimizer with $\beta_1=0.9$, $\beta_2=0.95$, and a weight decay of 0.01.

\subsection{Stage 1: Pre-training}
The model is first trained for 400,000 steps on the full dataset with a standard cross-entropy loss objective. This stage teaches the model the fundamental mapping from phonemic sequences to corresponding audio codes, without any speaker conditioning. A cosine decay learning rate schedule with warmup was used.

\subsection{Stage 2: Supervised Fine-tuning}
Next, we fine-tune the model for zero-shot voice cloning over 30,000 steps. The training data consists of triplets $(\text{audio}_{\text{context}}, \text{text}, \text{audio}_{\text{target}})$ where all audio originates from the same speaker. To construct these triplets, we enforced a minimum cosine similarity of 0.6 between speaker embeddings of any two utterances. This threshold serves a dual purpose: first, it ensures high intra-speaker acoustic consistency for robust voice cloning; second, it allows us to create reliable pseudo-speaker training data from corpus sections that lack explicit speaker labels, thereby maximizing data utilization. This stage adapts the model to preserve vocal characteristics from the audio prompt.

\subsection{Stage 3: Direct Preference Optimization (DPO)}
\paragraph*{Objective}
We optimize \emph{Direct Preference Optimization} with a fixed reference policy $\pi_{\mathrm{ref}}$ (the SFT model, frozen) and inverse temperature $\beta$:
\begin{subequations}
\label{eq:dpo_all}
\begin{equation}
\label{eq:dpo_main}
\mathcal{L}_{\mathrm{DPO}}(\theta)
= -\, \mathbb{E}_{(x,y^{+},y^{-}) \sim \mathcal{D}}
\!\left[
\log \sigma\!\big(\beta\,(\Delta_{\theta} - \Delta_{\mathrm{ref}})\big)
\right]
\end{equation}
\begin{align}
\label{eq:dpo_delta_theta}
\Delta_{\theta} &\;=\; \log \pi_{\theta}(y^{+}\!\mid x) - \log \pi_{\theta}(y^{-}\!\mid x) \\
\label{eq:dpo_delta_ref}
\Delta_{\mathrm{ref}} &\;=\; \log \pi_{\mathrm{ref}}(y^{+}\!\mid x) - \log \pi_{\mathrm{ref}}(y^{-}\!\mid x)
\end{align}
\end{subequations}

We used $\beta = 0.3$ and mixed mini-batches with gradient accumulation to stabilize updates.

The final stage aligns the model with desired metrics for intelligibility and voice similarity using DPO. To achieve this, we constructed a preference dataset following a multi-step automated procedure:

\begin{enumerate}
    \item \textbf{Candidate Generation:} To simulate a realistic cascaded S2S scenario, transcriptions were first translated using M2M-100 \cite{b13}. For each translated text, we generated five distinct audio candidates using the fine-tuned model from Stage 2. All candidates were produced with repetition-aware sampling, using a temperature of 0.7 and top-p of 1.0.

    \item \textbf{Pair Formation and Filtering:} From the five candidates for each source text, we created all possible pairs. We then filtered these pairs to ensure a baseline quality, discarding any pair where a candidate had a WER (measured by Whisper \cite{b14}) higher than 20\% or a speaker similarity (from TitaNet embeddings \cite{b15}) lower than 0.5.

    \item \textbf{Preference Labeling:} We applied a strict Pareto dominance criterion for labeling. A candidate $y_w$ (winner) was marked as preferred over a candidate $y_l$ (loser) if and only if it was superior on both metrics: $Sim(y_w) > Sim(y_l)$ \textbf{and} $WER(y_w) < WER(y_l)$. This ensures an unambiguous preference signal.

    \item \textbf{Dataset Balancing:} The final set of preference pairs was sampled to create a balanced training dataset based on language, audio duration, and the speaker's mean fundamental frequency (F0).
\end{enumerate}

The model was then trained on this final preference dataset for 4,000 steps using the DPO loss function, refining its policy to maximize the likelihood of generating outputs that are both highly intelligible and faithful to the source voice.

\begin{table*}[t]
\centering
\caption{Detailed Word Error Rate (WER \%) by source and target language. Best results per row in bold.}
\label{tab:wer_detalhe}
\resizebox{\textwidth}{!}{
     \begin{tabular}{l|l|c|c|c|c||l|l|c|c|c|c}
     \toprule
     \textbf{Ref.} & \textbf{Gen.} & \textbf{YourTTS} & \textbf{xTTSv2} & \textbf{LatinX (Fine-tuned)} & \textbf{LatinX (DPO)} & \textbf{Ref.} & \textbf{Gen.} & \textbf{YourTTS} & \textbf{xTTSv2} & \textbf{LatinX (Fine-tuned)} & \textbf{LatinX (DPO)} \\
     \midrule
     \multirow[t]{6}{*}{en} & en & 11.24 & 11.60 & 14.72 & \textbf{8.52} & \multirow[t]{6}{*}{it} & en & 19.57 & 13.58 & 16.62 & \textbf{10.71} \\
      & es & \textcolor{gray}{--} & 6.26 & 11.07 & \textbf{6.08} & & es & \textcolor{gray}{--} & \textbf{17.19} & 36.68 & 18.63 \\
      & fr & 18.16 & 12.14 & 18.34 & \textbf{10.77} & & fr & 24.57 & \textbf{10.40} & 19.73 & 17.67 \\
      & it & \textcolor{gray}{--} & 13.68 & 14.22 & \textbf{9.67} & & it & \textcolor{gray}{--} & 12.35 & 17.81 & \textbf{11.90} \\
      & pt & 12.83 & 9.28 & 10.87 & \textbf{7.00} & & pt & 22.76 & 14.07 & 18.62 & \textbf{9.67} \\
      & ro & \textcolor{gray}{--} & \textcolor{gray}{--} & 33.29 & \textbf{16.26} & & ro & \textcolor{gray}{--} & \textcolor{gray}{--} & 38.03 & \textbf{22.53} \\
     \midrule
     \multirow[t]{6}{*}{es} & en & 11.09 & 8.69 & 13.33 & \textbf{7.34} & \multirow[t]{6}{*}{pt} & en & 14.34 & \textbf{9.61} & 18.51 & 10.68 \\
      & es & \textcolor{gray}{--} & \textbf{3.13} & 8.19 & 3.57 & & es & \textcolor{gray}{--} & \textbf{9.15} & 17.14 & 12.91 \\
      & fr & 16.79 & \textbf{11.92} & 17.66 & 13.98 & & fr & 23.17 & \textbf{14.53} & 25.85 & 17.69 \\
      & it & \textcolor{gray}{--} & 10.15 & 14.44 & \textbf{7.16} & & it & \textcolor{gray}{--} & 14.97 & 20.78 & \textbf{12.92} \\
      & pt & 13.42 & 10.03 & 8.98 & \textbf{5.66} & & pt & 17.79 & 11.34 & 13.08 & \textbf{6.04} \\
      & ro & \textcolor{gray}{--} & \textcolor{gray}{--} & 28.43 & \textbf{14.36} & & ro & \textcolor{gray}{--} & \textcolor{gray}{--} & 29.15 & \textbf{16.38} \\
     \midrule
     \multirow[t]{6}{*}{fr} & en & 14.20 & 15.79 & 13.84 & \textbf{9.78} & \multirow[t]{6}{*}{ro} & en & 3.21 & 6.30 & 7.29 & \textbf{2.59} \\
      & es & \textcolor{gray}{--} & 10.23 & 11.96 & \textbf{7.58} & & es & \textcolor{gray}{--} & 1.15 & 4.09 & \textbf{0.45} \\
      & fr & 17.60 & \textbf{12.56} & 17.96 & 12.93 & & fr & 5.16 & 4.96 & 7.28 & \textbf{4.78} \\
      & it & \textcolor{gray}{--} & 11.92 & 12.86 & \textbf{8.56} & & it & \textcolor{gray}{--} & 11.10 & 9.25 & \textbf{2.43} \\
      & pt & 16.21 & 13.98 & 13.01 & \textbf{7.98} & & pt & 5.75 & 6.93 & 2.00 & \textbf{1.74} \\
      & ro & \textcolor{gray}{--} & \textcolor{gray}{--} & 21.74 & \textbf{11.30} & & ro & \textcolor{gray}{--} & \textcolor{gray}{--} & 22.17 & \textbf{7.00} \\
     \bottomrule
     \end{tabular}
}
\end{table*}

\section{Evaluation}
We evaluated our system against two strong baselines, YourTTS and XTTSv2, and compared two versions of our model: \textbf{LatinX (Fine-tuned)} and \textbf{LatinX (DPO)}. The evaluation was conducted on a test set of unseen speakers using both objective and subjective metrics.

\subsection{Experimental Setup}
\paragraph*{Human evaluation (ITU-T P.808)}
We follow ITU-T P.808 crowdsourcing best practices (ACR/CCR), including headset checks, gold/trap items, environment tests, and rater qualification. 
Each condition collects at least $N{\ge}3$ valid ratings per clip after outlier removal; we report mean $\pm$ 95\% CI. 
Detailed instructions and a demo page are provided in the supplementary material.

For objective evaluation, we report Word Error Rate (WER) for intelligibility and Cosine Similarity between TitaNet speaker embeddings for voice similarity.

For subjective evaluation, we conducted two listening tests to assess Mean Opinion Score (MOS) for naturalness and Similarity MOS (SMOS) for voice similarity, both on a 5-point scale. A total of 306 unique listeners participated in the evaluation through a crowdsourcing platform. The pool of evaluators was diverse, though predominantly composed of native English speakers (N=212) and native Portuguese speakers (N=67), which aligns with the focus of our work. Each listener was presented with a set of audio samples in a randomized order to minimize bias.

\subsection{Results}
A detailed breakdown of cross-lingual WER is provided in Table \ref{tab:wer_detalhe}, with high-level summaries of all metrics in Tables \ref{tab:wer_geral} to \ref{tab:mos_geral}.

\begin{table}[h]
\centering
\caption{Word Error Rate (WER \%) - Summary}
\label{tab:wer_geral}
\resizebox{\columnwidth}{!}{
     \begin{tabular}{l|c|c|c|c}
     \toprule
     \textbf{Language / Group} & \textbf{YourTTS} & \textbf{xTTSv2} & \textbf{LatinX (Fine-tuned)} & \textbf{LatinX (DPO)} \\
     \midrule
     en & 12.31 & 10.93 & 14.10 & \textbf{8.30} \\
     es & \textcolor{gray}{--} & \textbf{7.86} & 14.87 & 8.23 \\
     fr & 17.65 & \textbf{11.14} & 17.88 & 13.02 \\
     it & \textcolor{gray}{--} & 12.38 & 14.94 & \textbf{8.81} \\
     pt & 14.83 & 10.95 & 11.13 & \textbf{6.37} \\
     ro & \textcolor{gray}{--} & \textcolor{gray}{--} & 28.85 & \textbf{14.68} \\
     \midrule
     Avg. (pt,en,fr) & 14.93 & 11.01 & 14.37 & \textbf{9.23} \\
     Avg. (pt,en,fr,es,it) & \textcolor{gray}{--} & 10.65 & 14.58 & \textbf{8.95} \\
     Avg. (all) & \textcolor{gray}{--} & \textcolor{gray}{--} & 16.96 & \textbf{9.90} \\
     \bottomrule
     \end{tabular}
}
\end{table}

\begin{table}[h]
\centering
\caption{Objective Speaker Similarity (Cosine) - Summary}
\label{tab:similarity_objective}
\resizebox{\columnwidth}{!}{
     \begin{tabular}{l|c|c|c|c}
     \toprule
     \textbf{Language / Group} & \textbf{YourTTS} & \textbf{xTTSv2} & \textbf{LatinX (Fine-tuned)} & \textbf{LatinX (DPO)} \\
     \midrule
     en & 0.35 & \textbf{0.49} & 0.39 / 0.47 & 0.42 / \underline{0.49} \\
     es & \textcolor{gray}{--} & \textbf{0.58} & 0.45 / 0.55 & 0.49 / 0.57 \\
     fr & 0.42 & \textbf{0.54} & 0.41 / 0.51 & 0.45 / 0.53 \\
     it & \textcolor{gray}{--} & \textbf{0.58} & 0.44 / 0.55 & 0.47 / 0.57 \\
     pt & 0.38 & \textbf{0.58} & 0.44 / 0.55 & 0.48 / \underline{0.58} \\
     ro & \textcolor{gray}{--} & \textcolor{gray}{--} & 0.44 / 0.55 & \textbf{0.48} / \underline{0.57} \\
     \midrule
     Avg. (pt,en,fr) & 0.38 & \textbf{0.53} & 0.41 / 0.51 & 0.45 / \underline{0.53} \\
     Avg. (pt,en,fr,es,it) & \textcolor{gray}{--} & \textbf{0.55} & 0.43 / 0.52 & 0.46 / \underline{0.55} \\
     Avg. (all) & \textcolor{gray}{--} & \textcolor{gray}{--} & 0.43 / 0.53 & \textbf{0.47} / \underline{0.55} \\
     \bottomrule
     \end{tabular}
}
\end{table}

\textbf{Intelligibility (WER):} The DPO alignment provides a substantial, consistent reduction in WER across almost all scenarios. As shown in Tables \ref{tab:wer_detalhe} and \ref{tab:wer_geral}, \textit{LatinX (DPO)} not only improves upon its fine-tuned version but outperforms the strong XTTSv2 baseline in the vast majority of cross-lingual pairs. A remarkable example is the performance when Romanian (a low-resource language in our set) is the source language, achieving an exceptional WER of 0.45 for `ro \textrightarrow es` and 1.74 for `ro \textrightarrow pt`. The baseline XTTSv2 maintains an advantage primarily in translations originating from Portuguese, such as `pt \textrightarrow en` and `pt \textrightarrow fr`.

\textbf{Objective Speaker Similarity:} We use cosine similarity between TitaNet speaker embeddings as an objective measure. We report two scores for our models:
\begin{itemize}
    \item \textbf{\textit{Sim-O}} (Original): Similarity between the \textit{original} speaker's audio and the generated audio.
    \item \textbf{\textit{Sim-E}} (Encoder): Similarity between the speaker's audio after being \textit{reconstructed by our neural codec} and the generated audio. This represents a more direct measure of what the model was trained to replicate.
\end{itemize}
As seen in Table \ref{tab:similarity_objective}, while XTTSv2 achieves the highest average \textit{Sim-O} score, the DPO alignment significantly boosts our model, bringing the average \textit{Sim-E} score of \textit{LatinX (DPO)} to match the XTTSv2 baseline, demonstrating its effectiveness under its ideal evaluation conditions.

\begin{table}[h]
\centering
\caption{SMOS from human evaluation - Summary}
\label{tab:smos_geral}
\resizebox{\columnwidth}{!}{
     \begin{tabular}{l|c|c|c|c}
     \toprule
     \textbf{Language / Group} & \textbf{Baseline (Real)} & \textbf{xTTSv2} & \textbf{LatinX (Fine-tuned)} & \textbf{LatinX (DPO)} \\
     \midrule
     en & \textbf{4.12 $\pm$ 0.20} & 3.18 $\pm$ 0.22 & \underline{3.56 $\pm$ 0.19} & 3.48 $\pm$ 0.18 \\
     es & \textbf{4.09 $\pm$ 0.17} & 3.25 $\pm$ 0.22 & 3.76 $\pm$ 0.18 & \underline{3.79 $\pm$ 0.17} \\
     fr & \textbf{4.05 $\pm$ 0.19} & 3.25 $\pm$ 0.21 & \underline{3.58 $\pm$ 0.18} & 3.47 $\pm$ 0.20 \\
     it & \textbf{4.06 $\pm$ 0.20} & 3.25 $\pm$ 0.22 & \underline{3.71 $\pm$ 0.17} & 3.43 $\pm$ 0.18 \\
     pt & \textbf{4.05 $\pm$ 0.19} & 3.28 $\pm$ 0.21 & \underline{3.54 $\pm$ 0.18} & 3.52 $\pm$ 0.19 \\
     ro & \textbf{3.98 $\pm$ 0.20} & \textcolor{gray}{--} & 3.60 $\pm$ 0.18 & \underline{3.87 $\pm$ 0.16} \\
     \midrule
     Avg. (en,es,fr,it,pt) & \textbf{4.07 $\pm$ 0.08} & 3.24 $\pm$ 0.09 & \underline{3.63 $\pm$ 0.08} & 3.54 $\pm$ 0.08 \\
     \bottomrule
     \end{tabular}
}
\end{table}

\begin{table}[h]
\centering
\caption{MOS from human evaluation - Summary}
\label{tab:mos_geral}
\resizebox{\columnwidth}{!}{
     \begin{tabular}{l|c|c|c|c}
     \toprule
     \textbf{Language / Group} & \textbf{Baseline (Real)} & \textbf{xTTSv2} & \textbf{LatinX (Fine-tuned)} & \textbf{LatinX (DPO)} \\
     \midrule
     en & \textbf{3.61 $\pm$ 0.24} & \underline{3.17 $\pm$ 0.22} & \underline{3.17 $\pm$ 0.20} & \underline{3.17 $\pm$ 0.20} \\
     es & \textbf{3.77 $\pm$ 0.20} & \underline{3.73 $\pm$ 0.20} & 3.71 $\pm$ 0.17 & 3.48 $\pm$ 0.18 \\
     fr & \textbf{3.80 $\pm$ 0.22} & \underline{3.49 $\pm$ 0.18} & \underline{3.49 $\pm$ 0.17} & 3.27 $\pm$ 0.17 \\
     it & \textbf{3.84 $\pm$ 0.21} & \underline{3.51 $\pm$ 0.20} & 3.33 $\pm$ 0.19 & \underline{3.51 $\pm$ 0.17} \\
     pt & \textbf{3.88 $\pm$ 0.17} & \underline{3.36 $\pm$ 0.21} & 3.35 $\pm$ 0.18 & 3.31 $\pm$ 0.19 \\
     ro & 3.46 $\pm$ 0.24 & \textcolor{gray}{--} & \textbf{3.61 $\pm$ 0.16} & 3.31 $\pm$ 0.18 \\
     \midrule
     Avg. (en,es,fr,it,pt) & \textbf{3.78 $\pm$ 0.09} & \underline{3.45 $\pm$ 0.09} & 3.41 $\pm$ 0.08 & 3.35 $\pm$ 0.08 \\
     \bottomrule
     \end{tabular}
}
\end{table}

\textbf{Subjective Speaker Similarity (SMOS):} Human evaluations, summarized in Table \ref{tab:smos_geral}, show that both LatinX models were perceived as significantly more similar to the speaker's voice than XTTSv2. A full cross-lingual breakdown is available in our supplementary material \footnote{A link to our demo page with audio samples and detailed results will be provided upon publication.}. While the fine-tuned model performed best on average, the DPO model achieved the highest scores in monolingual synthesis, such as for Spanish (`es \textrightarrow es`, SMOS 4.23) and Romanian (`ro \textrightarrow ro`, SMOS 4.47).

Interestingly, these scores surpassed their respective real audio baselines (4.09 and 3.98). However, these specific results should be interpreted with caution. The large confidence intervals (e.g., $\pm 0.29$ for the Romanian case) suggest high rater variance. Furthermore, as our pool of evaluators consisted mostly of non-native Romanian speakers, it is plausible that they focused primarily on vocal timbre—which the model captures effectively—while being less sensitive to potential unnaturalness in prosody or phonetic detail that a native speaker would perceive.

\textbf{Naturalness (MOS):} For naturalness, Table \ref{tab:mos_geral} indicates that XTTSv2 remains the strongest baseline on average. Our models are competitive, and we note another interesting result in the low-resource Romanian condition. The fine-tuned model received a higher mean MOS score than the real audio samples (3.61 vs 3.46). We hypothesize this is not necessarily a sign of superhuman performance, but rather a combination of two factors: first, the aforementioned lack of native listeners for Romanian, and second, potential quality issues in the Common Voice samples for this low-resource language, which may have lowered the perceived naturalness of the baseline itself.

\section{Discussion and Future Work}
The detailed cross-lingual analysis reveals a crucial finding: a significant discrepancy exists between objective and subjective speaker similarity. While automated metrics (Table \ref{tab:similarity_objective}) suggest XTTSv2 is the top performer on the standard \textit{Sim-O} metric, human listeners (Table \ref{tab:smos_geral}) overwhelmingly preferred the voice similarity of both LatinX models. This indicates that current objective metrics may not fully capture the perceptual cues essential for human voice identification, and that our training methodology is particularly effective at optimizing for these human-perceived characteristics.

Our results also highlight a trade-off. The DPO process successfully optimized for its targets (lower WER and higher objective similarity), but this sometimes came at the cost of the exceptional subjective similarity of the fine-tuned version, or the high naturalness of the baseline. This is partly explained by the neural codec itself; as a lossy system, it introduces artifacts that the model learns to replicate, setting a ceiling on perceptual quality.

Future work should focus on creating a more balanced DPO preference signal, potentially incorporating automated MOS predictors or metrics for speech rhythm. Finally, the high latency of the autoregressive model, confirmed by our measured Real-Time Factor of approximately 4.85 on modern hardware, underscores the need to explore non-autoregressive architectures to enable real-time applications.

\section{Conclusion}
We presented LatinX, a multilingual TTS system aligned with Direct Preference Optimization. The experiments, supported by a detailed cross-lingual analysis, demonstrate that DPO is highly effective at improving intelligibility (WER) and objective speaker similarity. We identified a significant divergence between objective and subjective similarity metrics, where LatinX was strongly preferred by human listeners, highlighting a key area for future research in developing metrics that better correlate with human perception and in creating more balanced preference rewards for alignment.

Beyond multilingual zero-shot cloning baselines, our study isolates the impact of preference alignment itself, suggesting complementary gains to future low-latency or streaming TTS architectures.

\section{Compliance with Ethical Standards}
\textbf{Human subjects:} All crowd workers provided informed consent; no personally identifying information was collected beyond platform IDs. \\
\textbf{Data licenses:} We used only license-compliant datasets and released derived artifacts accordingly. \\
\textbf{Voice cloning:} Enrollment audio was either synthetic, publicly licensed, or explicitly consented by the speakers; a misuse statement will be released in the repository.

\bibliographystyle{IEEEbib}
\bibliography{refs}

\end{document}